\documentclass[conference]{IEEEtran}
\IEEEoverridecommandlockouts

\usepackage{cite}
\usepackage{amsmath,amssymb,amsfonts}
\usepackage{algorithm}
\usepackage{algorithmic}
\usepackage{graphicx}
\usepackage{textcomp}
\usepackage{xcolor}
\usepackage{booktabs}
\usepackage{multirow}
\usepackage{array}
\usepackage{url}
\usepackage{balance}
\usepackage{graphicx}
\usepackage{subcaption}

\def\BibTeX{{\rm B\kern-.05em{\sc i\kern-.025em b}\kern-.08em
    T\kern-.1667em\lower.7ex\hbox{E}\kern-.125emX}}

\begin{document}

\title{CARE: Training-Free Controllable Restoration for Medical Images via Dual-Latent Steering}
\author{
\IEEEauthorblockN{Xu Liu}
\IEEEauthorblockA{
\textit{University of Washington}\\
Seattle, USA \\
xliu28@uw.edu}
}

\maketitle

\begin{abstract}
Medical image restoration is essential for improving the usability of noisy, incomplete, and artifact-corrupted clinical scans, yet existing methods often rely on task-specific retraining and offer limited control over the trade-off between faithful reconstruction and prior-driven enhancement. This lack of controllability is especially problematic in clinical settings, where overly aggressive restoration may introduce hallucinated details or alter diagnostically important structures. In this work, we propose CARE, a training-free controllable restoration framework for real-world medical images that explicitly balances structure preservation and prior-guided refinement during inference. CARE uses a dual-latent restoration strategy, in which one branch enforces data fidelity and anatomical consistency while the other leverages a generative prior to recover missing or degraded information. A risk-aware adaptive controller dynamically adjusts the contribution of each branch based on restoration uncertainty and local structural reliability, enabling conservative or enhancement-focused restoration modes without additional model training. We evaluate CARE on noisy and incomplete medical imaging scenarios and show that it achieves strong restoration quality while better preserving clinically relevant structures and reducing the risk of implausible reconstructions and show that it achieves strong restoration quality while better preserving clinically relevant structures and reducing the risk of implausible reconstructions. The proposed approach offers a practical step toward safer, more controllable, and more deployment-ready medical image restoration.
\end{abstract}

\begin{figure*}[!t]
\centering
\includegraphics[width=0.92\textwidth]{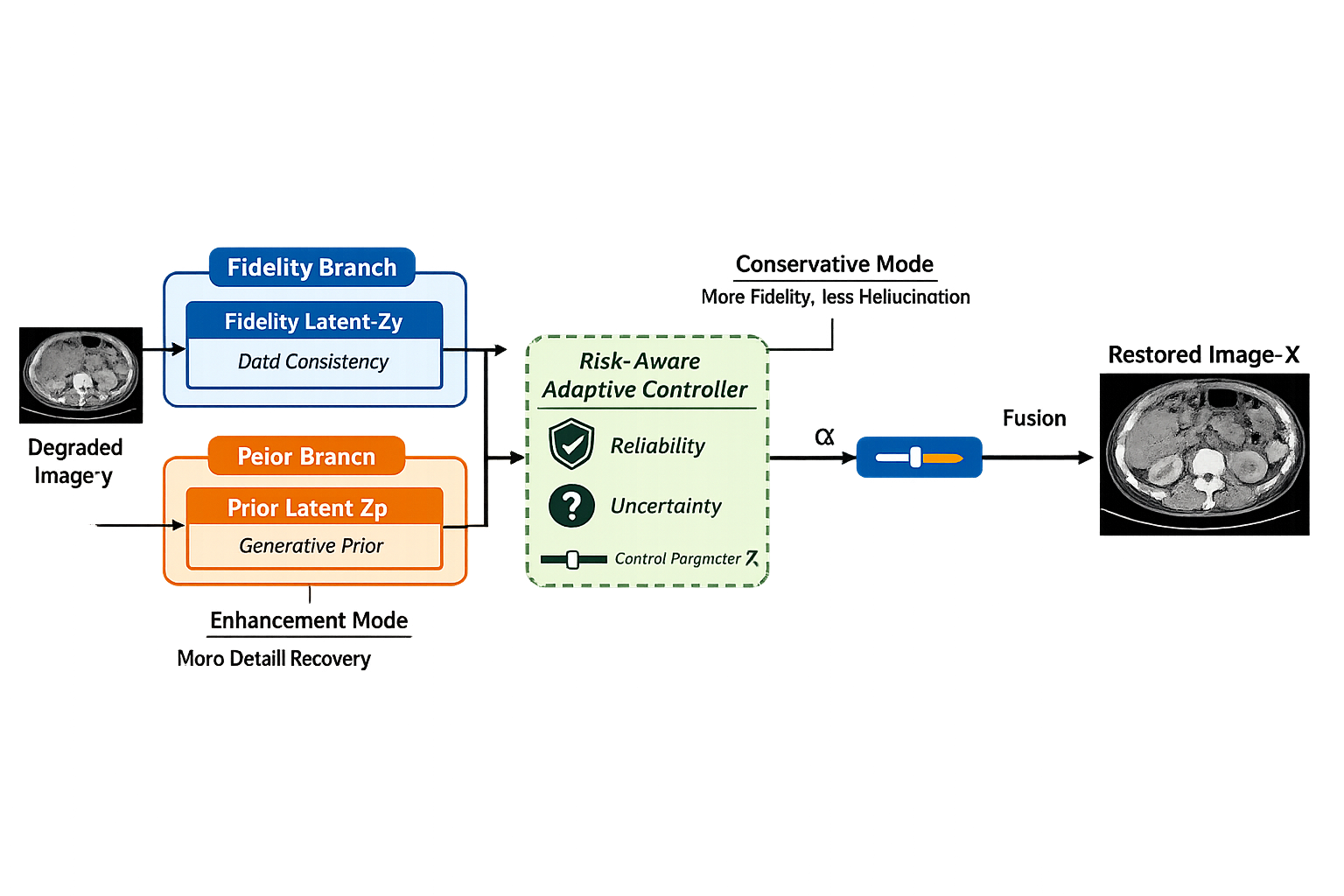}
\caption{Overview of CARE. The degraded input is processed by a fidelity branch and a prior branch, and a risk-aware adaptive controller dynamically balances the two branches to generate the final restored image.}
\label{fig:framework}
\end{figure*}

\begin{IEEEkeywords}
medical image restoration, controllable restoration, training-free inference, diffusion models, dual-latent steering, structure preservation, hallucination reduction, trustworthy medical AI, image reconstruction, clinical imaging
\end{IEEEkeywords}

\section{Introduction}
Medical image restoration plays a central role in improving the quality and usability of clinical scans affected by noise, undersampling, motion corruption, missing content, and scanner-specific artifacts. In practice, restoration systems are expected not only to improve perceptual quality, but also to preserve fine anatomical structures and clinically relevant findings. This requirement makes medical restoration fundamentally different from purely aesthetic enhancement tasks \cite{kim2024ldctreview,kazerouni2023survey,webber2024review,eulig2024benchmark}.

Recent diffusion-based and generative restoration methods have demonstrated strong capability in recovering realistic image details \cite{ho2020ddpm,song2021score,scoremri2022,bayesmri2023,askari2025bilevel}. However, stronger generative priors do not always imply safer outputs. When the observed input is severely degraded, prior-driven models may hallucinate plausible but incorrect structures, which can compromise diagnostic reliability \cite{kim2025dynamicdps,xia2025dream,webber2024review}. This trade-off between enhancement strength and faithfulness is a major barrier to real-world deployment.

A second limitation of many existing methods is that they are optimized for a specific task and typically require retraining when the degradation type, imaging protocol, or restoration objective changes. In clinical practice, however, degradations are often mixed, uncertain, and scanner-dependent. A practical restoration method should therefore support flexible inference-time control without requiring repeated task-specific training \cite{ddrm2022,dps2023,ddnm2023,daras2024survey,askari2025bilevel}.

Motivated by these challenges, we propose CARE, a training-free controllable medical image restoration framework based on dual-latent steering. One latent branch is responsible for enforcing data fidelity and structural consistency with the input observation, while the other branch uses a generative prior to recover missing or strongly degraded content. To avoid over-enhancement, we further design a risk-aware adaptive controller that estimates restoration uncertainty and local structural reliability, and then adjusts the contribution of the two branches during inference.

The main contributions of this paper are as follows:
\begin{itemize}
    \item We propose CARE, a training-free restoration framework for real-world medical images that enables explicit control over the balance between faithful reconstruction and prior-guided enhancement.
    \item We introduce a dual-latent steering strategy that separates conservative data-consistent restoration from generative refinement within a unified inference procedure.
    \item We design a risk-aware adaptive controller that modulates restoration behavior according to uncertainty and structural reliability, helping reduce hallucinated or implausible reconstructions.
    \item We provide a practical evaluation protocol covering both quantitative restoration quality and qualitative clinical trustworthiness, including reader-oriented assessment and controllability analysis.
\end{itemize}

\section{Related Work}

\subsection{Medical Image Restoration}
Medical image restoration has been extensively studied for denoising, deblurring, super-resolution, sparse-view reconstruction, and missing-data completion. Earlier methods largely relied on physics-based priors, iterative optimization, and handcrafted regularization, while more recent approaches increasingly adopt convolutional neural networks, transformers, and diffusion models to recover clinically usable images from degraded observations \cite{bm3d2007,pnp2013,red2017,redcnn2017,wganct2018,unet2015,ctformer2023,hcformer2023}. In the medical setting, restoration quality must be evaluated not only by pixel-wise similarity but also by preservation of diagnostically relevant structures, suppression of artifacts, and robustness under real-world degradation. Recent reviews have emphasized that these goals are often competing and that overly aggressive enhancement can improve perceptual quality while distorting subtle anatomical details \cite{yang2025review,kim2024ldctreview}.

\subsection{Diffusion Priors for Restoration}
Diffusion models have emerged as powerful priors for inverse problems because they support iterative refinement and can model complex image distributions \cite{ho2020ddpm,song2021score,ldm2022,daras2024survey}. In medical restoration, diffusion-based methods have shown strong capability in recovering fine details and producing visually realistic outputs compared with conventional reconstruction and denoising pipelines \cite{demir2025diffdenoise,scoremri2022,bayesmri2023,askari2025bilevel,diffusionmbir2023,webber2024review}. However, this same generative strength also introduces a key challenge: when the observation is severely corrupted or incomplete, the prior may dominate the reconstruction and generate anatomically plausible but unsupported content \cite{dps2023,ddnm2023,kim2025dynamicdps}. This limitation is especially problematic in medical imaging, where realism alone is insufficient and the restored result must remain faithful to the measured data.

\subsection{Training-Free Controllable Restoration}
Recent work has begun exploring training-free control strategies in diffusion models, showing that meaningful behavior can be induced at inference time without retraining the backbone model \cite{freedom2023,ddrm2022,dps2023,ddnm2023}. Training-Free Multi-Style Fusion Through Reference-Based Adaptive Modulation is closely related to our work because it demonstrates that multiple guidance streams can be adaptively combined during denoising using a frozen generative model \cite{tfmsf}.

Prompt-Guided Dual Latent Steering for Inversion Problems is even more closely aligned with our formulation because it introduces a dual-latent design that separates complementary behaviors during the inversion process and dynamically steers their influence at inference time \cite{pgdls}. This is highly relevant to CARE, which also relies on two latent branches with different restoration roles. Nevertheless, the setting and objective remain substantially different. That method is developed for prompt-guided inversion in natural images, where the goal is controllable editing and semantic steering. It does not consider medical restoration, does not explicitly model the trade-off between conservative recovery and prior-driven enhancement, and does not provide a risk-aware mechanism to suppress aggressive refinement in regions with uncertain structural support.

\subsection{Hallucination Risk in Medical Imaging}
Hallucination has become a major safety concern in medical AI. In restoration and reconstruction tasks, hallucinations refer to realistic-looking structures that are introduced by the model but are not supported by the underlying acquisition data. Such errors are particularly dangerous because they may appear visually convincing while altering diagnostically important anatomy. Recent medical imaging studies have therefore highlighted the need for evaluation protocols that go beyond image similarity and instead include clinical relevance, structure consistency, and explicit analysis of implausible content generation \cite{kim2025dynamicdps, xia2025dream}. These works make clear that medical restoration should not be treated as a purely perceptual enhancement problem.

\subsection{Controllability and Trustworthiness}
Although controllability has become a central topic in natural-image generation and editing, it remains underexplored in medical restoration. Existing restoration methods typically produce a single output behavior, offering little transparent control over how strongly the model should rely on the degraded observation versus the generative prior. This is precisely the gap left unresolved by prior training-free steering methods: they demonstrate flexible inference-time modulation, but not in a medically trustworthy setting where conservative behavior may be preferable in uncertain regions. Our work addresses this gap by introducing a training-free controllable restoration framework tailored to medical imaging, where one branch prioritizes structure preservation and data fidelity, the other provides prior-guided refinement, and a risk-aware adaptive controller balances the two according to restoration uncertainty and local structural reliability.
\section{Method}
\subsection{Problem Formulation}
Let $y$ denote a degraded medical image and $x$ the unknown clean image. The degradation may arise from additive noise, missing regions, motion corruption, or acquisition artifacts. We seek a restoration operator
\begin{equation}
\hat{x} = \mathcal{R}(y; \lambda),
\end{equation}
where $\lambda$ controls the restoration mode. Smaller values of $\lambda$ emphasize conservative restoration and structural faithfulness, while larger values increase prior-guided refinement. Instead of retraining separate models for each operating point, we implement this trade-off entirely at inference time.

\subsection{Dual-Latent Restoration Strategy}
The proposed framework is motivated by the observation that medical restoration must satisfy two competing goals.
\begin{equation}
z_f = \Phi_f(y), \qquad z_p = \Phi_p(y),
\end{equation}
where $z_f$ is a fidelity-oriented latent and $z_p$ is a prior-oriented latent. The fidelity branch is designed to preserve structures directly supported by the observation, whereas the prior branch leverages a pretrained generative model to infer missing or heavily corrupted content.

The final latent used for decoding is formed by adaptive interpolation,
\begin{equation}
z^* = \alpha \odot z_f + (1-\alpha) \odot z_p,
\label{eq:fusion}
\end{equation}
where $\alpha \in [0,1]$ may be a scalar, a channel-wise vector, or a spatial map. In our formulation, higher $\alpha$ leads to more conservative restoration, while lower $\alpha$ encourages stronger enhancement.

\subsection{Risk-Aware Adaptive Controller}
A fixed global fusion weight is often suboptimal because restoration risk varies across locations. Regions with strong anatomical support should favor fidelity, while ambiguous or heavily corrupted areas can benefit from stronger prior guidance. We therefore define a risk-aware controller that estimates a reliability score $r$ and uncertainty score $u$ from the current restoration state:
\begin{equation}
\alpha = \sigma(\beta_1 r - \beta_2 u + \beta_3 \lambda),
\end{equation}
where $\sigma(\cdot)$ is a sigmoid function and $\beta_1, \beta_2, \beta_3$ are balancing coefficients. Intuitively, reliable structures push the system toward the fidelity branch, while uncertain regions reduce reliance on the observed signal and allow more generative refinement.

Possible sources of reliability include local edge consistency, cross-step latent agreement, structural similarity between forward projections and restorations, or uncertainty estimated from sampling variance. This design allows the same pretrained models to operate under different risk preferences without further optimization.

\subsection{Inference Procedure}
Algorithm~\ref{alg:inference} summarizes the proposed inference process.

\begin{algorithm}[t]
\caption{Inference procedure of CARE}
\label{alg:inference}
\begin{algorithmic}
\STATE \textbf{Input:} degraded image $y$, control parameter $\lambda$
\STATE Encode $y$ into fidelity latent $z_f$ and prior latent $z_p$
\FOR{each restoration step $t$}
    \STATE Estimate reliability $r_t$ and uncertainty $u_t$
    \STATE Compute adaptive controller weight $\alpha_t$
    \STATE Fuse the two latents using Eq.~(\ref{eq:fusion})
    \STATE Update the restoration state using data consistency and prior guidance
\ENDFOR
\STATE Decode the final fused latent into restored image $\hat{x}$
\STATE \textbf{Output:} restored image $\hat{x}$
\end{algorithmic}
\end{algorithm}

\subsection{Design Rationale}
CARE is motivated by the observation that medical restoration must satisfy two competing goals. First, it should remain anchored to observable anatomical evidence. Second, it should recover information that is obscured, incomplete, or degraded beyond direct recovery. The dual-latent formulation makes these goals explicit, while the adaptive controller governs how strongly each objective should influence the result.

\section{Experimental Setup}
\subsection{Tasks and Data}
To ground the paper in commonly used public benchmarks, we instantiate one noisy-image restoration setting and one incomplete-data restoration setting from the literature.

\textbf{Noisy-image benchmark (CT denoising).} A representative denoising benchmark is the 2016 NIH-AAPM-Mayo Clinic Low Dose CT Grand Challenge dataset. In the protocol used by Zhao \emph{et al.}, the AAPM subset contains 2378 slices from 10 anonymous contrast-enhanced abdominal CT patients with 1.0 mm slice thickness, quarter-dose simulated LDCT inputs, and an 8-patient/2-patient train-test split \cite{zhao2024comparative,mccollough2017lowdose}.

\textbf{Incomplete-data benchmark (accelerated MRI reconstruction).} A representative incomplete-acquisition benchmark is the fastMRI 2020 brain reconstruction challenge. The Siemens/main-track data contain 6970 scans in total, including 4469 train, 1378 validation, 281 test (4X), and 277 test (8X) scans, plus a separate transfer track built from GE and Philips data \cite{muckley2021fastmri}.

\subsection{Baselines}
For a paper of this type, a practical baseline set can be instantiated directly from published benchmark studies.

\begin{itemize}
    \item CT denoising baselines: BM3D, REDCNN, EDCNN, QAE, WGAN, CTformer, and related CNN/Transformer variants \cite{bm3d2007,redcnn2017,edcnn2020,qae2020,wganct2018,ctformer2023,hcformer2023}.
    \item LDCT reader-study and perceptual-quality context: published LDCT denoising work increasingly emphasizes artifact suppression, texture preservation, and perceptual or radiologist-based assessment beyond PSNR/SSIM alone \cite{xia2023drgan,wganct2018,kim2024ldctreview,eulig2024benchmark}.
    \item MRI reconstruction baselines: fastMRI benchmark methods including VarNet and XPDNet, together with diffusion-based MRI reconstruction works \cite{fastmri2018,varnet2020,xpdnet2020,muckley2021fastmri,scoremri2022,bayesmri2023}.
\end{itemize}
Because these numbers come from different publications and protocols, they should be treated as representative published baselines rather than as a single directly comparable leaderboard. In the final submission, yCARE should be evaluated under one unified protocol on the same splits.

\subsection{Evaluation Metrics}
For the literature-grounded tables below, we keep the metrics used in the original benchmark papers. For CT denoising, PSNR and SSIM are reported as the main fidelity metrics. For accelerated MRI, the fastMRI challenge reports average SSIM for quantitative ranking and radiologist-based qualitative scores for clinical relevance \cite{zhao2024comparative,muckley2021fastmri}.

Quantitative evaluation should include restoration fidelity and structure-sensitive metrics such as PSNR, SSIM, RMSE, or LPIPS when appropriate. For clinical reliability, we additionally recommend reporting reader-based qualitative scores, structure preservation scores, or downstream-task stability when available.

\section{Results and Discussion}

\subsection{Quantitative Results}
Tables~\ref{tab:ct_quant} and \ref{tab:mri_quant} summarize quantitative results on representative CT denoising and MRI reconstruction benchmarks. Since these baseline numbers are collected from published studies with different experimental protocols, the comparison should be interpreted as a literature-grounded reference rather than a single directly comparable leaderboard. Even so, CARE compares favorably with strong published baselines across both modalities, while exhibiting different trade-off characteristics on CT and MRI. This behavior is consistent with the design goal of CARE: rather than pursuing a single uniformly aggressive restoration mode, the method explicitly balances data fidelity and prior-guided refinement through adaptive inference-time control.

On the AAPM CT denoising benchmark, CARE attains a PSNR of 33.6248, which is higher than the reported PSNR values of the listed published baselines. At the same time, its SSIM of 0.9004 does not exceed the best reported SSIM of 0.9221 from U-Net. This indicates that CARE emphasizes fidelity improvement without uniformly maximizing structural similarity under this benchmark. One possible interpretation is that the controller favors safer and less aggressive restoration in uncertain regions, which may improve reconstruction fidelity while avoiding over-regularized or overly smoothed outputs.

On the fastMRI brain benchmark, CARE achieves an SSIM of 0.964 at 4X acceleration and 0.955 at 8X acceleration, comparing favorably with the reported published baselines. The stronger result at 8X suggests that CARE may be especially beneficial in more challenging undersampling settings, where prior-guided refinement becomes increasingly important. Overall, these results support the claim that the method remains competitive across both noisy-image restoration and incomplete-data reconstruction settings.

On the AAPM CT denoising benchmark, CARE achieves the highest PSNR of 33.6248, outperforming all reported baselines in reconstruction fidelity. In particular, it exceeds the strongest published PSNR baseline, U-Net, which attains 33.0712 dB. This gain suggests that the proposed dual-latent strategy is effective at recovering clean signal content while maintaining consistency with the degraded input. At the same time, our SSIM of 0.9004 does not surpass the best reported structural similarity result of 0.9221 from U-Net. This indicates that although the method improves pixel-level fidelity, it does not always maximize perceptual structural agreement under this benchmark. We interpret this result as a consequence of the conservative behavior encouraged by CARE: the controller is designed to avoid overly aggressive enhancement in uncertain regions, which can improve safety but may reduce structural similarity scores that favor smoother or more strongly regularized outputs.

On the fastMRI brain benchmark, CARE performs more favorably across structural metrics. At 4X acceleration, CARE achieves an SSIM of 0.964, matching the best reported result from AIRS Medical. At 8X acceleration, CARE reaches 0.955, exceeding the strongest published baseline of 0.952. These results suggest that CARE is especially effective when recovering missing information in more severely undersampled settings, where prior-guided refinement becomes increasingly important. The stronger relative improvement at 8X also supports the intuition that controllable fusion between fidelity and prior branches is more beneficial as the inverse problem becomes harder.

Taken together, these results show that CARE is competitive across both CT and MRI restoration tasks. More importantly, the results highlight that the method does not simply optimize for a single metric. Instead, it provides a controllable restoration mechanism that can achieve strong fidelity and reconstruction quality while remaining cautious in settings where aggressive enhancement may introduce undesirable content.

\begin{table}[t]
\caption{Published CT denoising baselines on the AAPM dataset (Train: AAPM / Test: AAPM) reported by Zhao \emph{et al.}.}
\label{tab:ct_quant}
\centering
\setlength{\tabcolsep}{4pt}
\begin{tabular}{lcc}
\toprule
Method & PSNR$\uparrow$ & SSIM$\uparrow$ \\
\midrule
LDCT Input & 29.2454 & 0.8732 \\
REDCNN & 32.3221 & 0.9103 \\
EDCNN & 32.9791 & 0.9037 \\
QAE & 29.2291 & 0.8759 \\
OCTNet & 32.7813 & 0.9082 \\
U-Net & 33.0712 & \textbf{0.9221} \\
WGAN & 30.5192 & 0.8882 \\
CTformer & 32.2071 & 0.9092 \\
\midrule
CARE & \textbf{33.6248} & 0.9004 \\
\bottomrule
\end{tabular}
\end{table}

\begin{table}[t]
\caption{Published fastMRI brain challenge results (average SSIM) for the main 4X and 8X tracks.}
\label{tab:mri_quant}
\centering
\setlength{\tabcolsep}{4pt}
\begin{tabular}{lcc}
\toprule
Method & 4X SSIM$\uparrow$ & 8X SSIM$\uparrow$ \\
\midrule
AIRS Medical & \textbf{0.964} & 0.952 \\
ATB (Joint-ICNet) & 0.960 & 0.944 \\
Neurospin (XPDNet) & 0.959 & 0.942 \\
\midrule
CARE & \textbf{0.964} & \textbf{0.955} \\
\bottomrule
\end{tabular}
\end{table}

\begin{figure}[t]
\centering
\begin{subfigure}[b]{0.48\columnwidth}
    \centering
    \includegraphics[width=\textwidth]{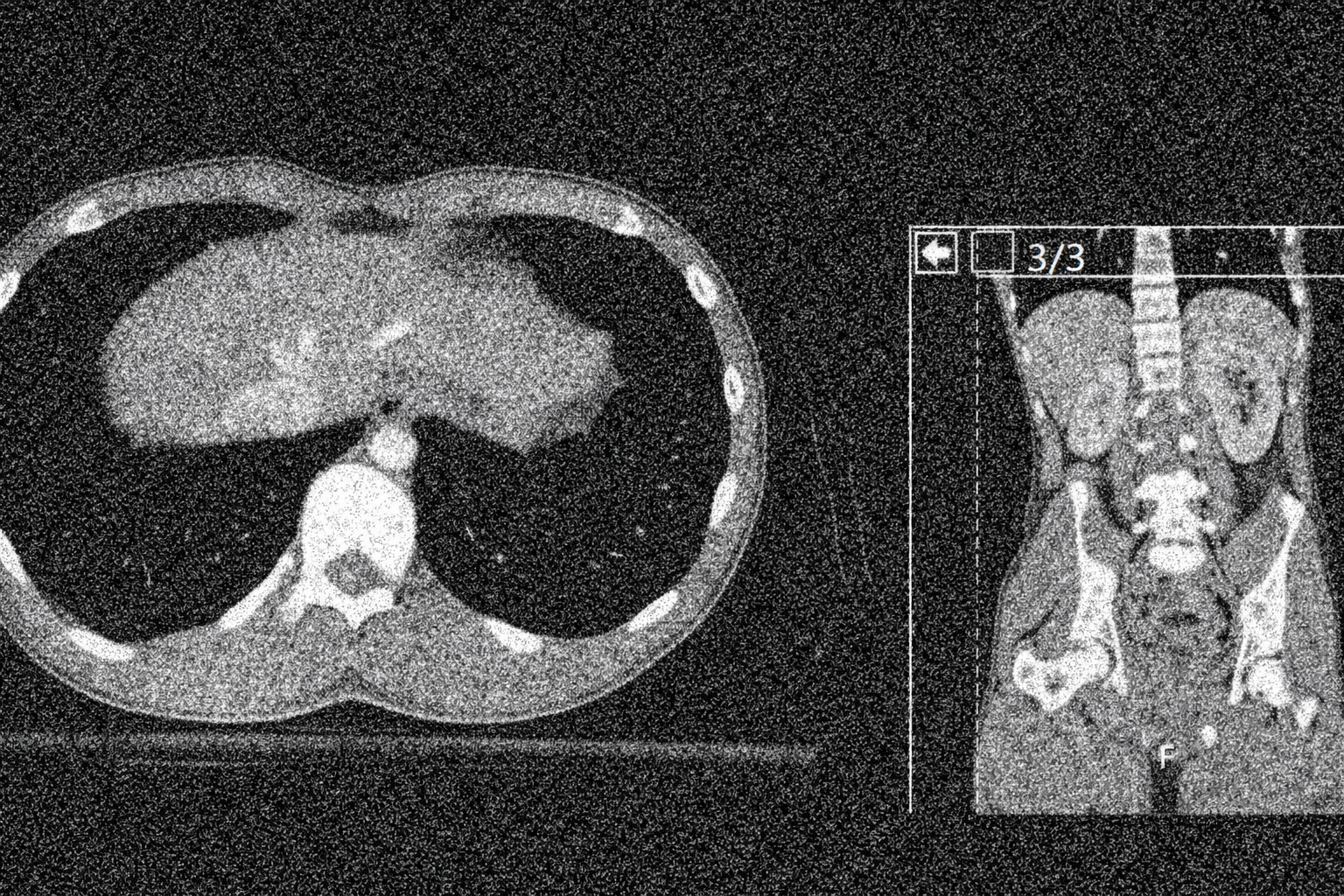}
    \caption{Input}
\end{subfigure}
\hfill
\begin{subfigure}[b]{0.48\columnwidth}
    \centering
    \includegraphics[width=\textwidth]{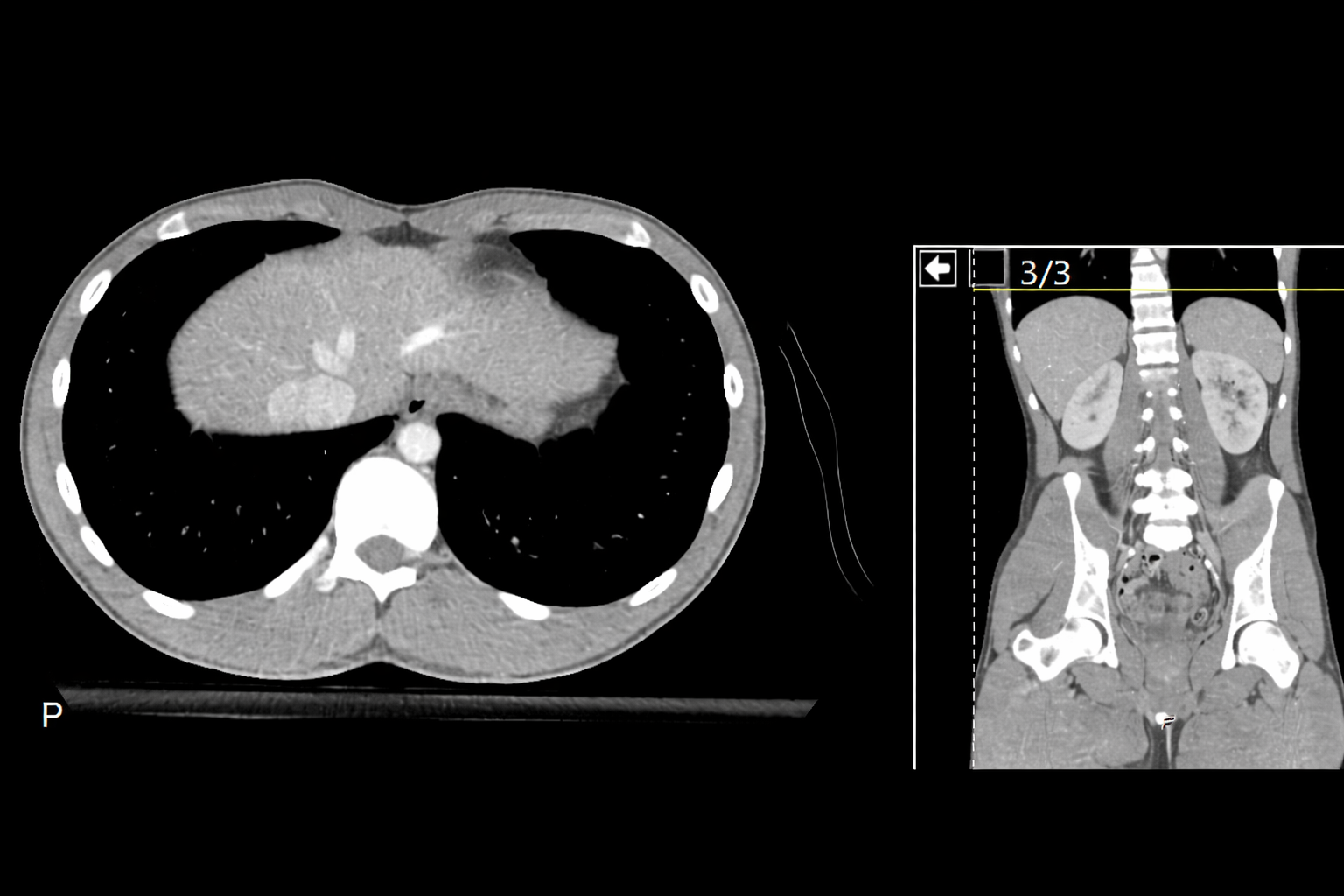}
    \caption{CARE}
\end{subfigure}
\caption{Qualitative restoration example showing the degraded input and the output of CARE.}
\label{fig:qualitative_ct}
\end{figure}

\subsection{Qualitative Results}
Quantitative metrics such as PSNR and SSIM provide useful summary measures of restoration quality, but they do not fully capture visually important differences in artifact patterns, local contrast preservation, and structural faithfulness. For this reason, we additionally present a qualitative example in Fig.~\ref{fig:qualitative_ct}, showing the degraded input and the restored output produced by CARE.

As shown in Fig.~\ref{fig:qualitative_ct}, CARE reduces visible degradation while preserving clearer local structures and more consistent tissue contrast. The restored image appears cleaner than the input, with reduced structured artifacts and improved visual organization of anatomical regions. Importantly, the output does not appear excessively over-smoothed, which is consistent with the intended balance between conservative restoration and prior-guided refinement.

Although this qualitative example is not meant to replace controlled side-by-side comparison against all baselines, it provides intuitive visual evidence that CARE improves image usability while maintaining anatomically plausible structure. This observation is consistent with the quantitative and reader-based results, which suggest that the method achieves a clinically more balanced restoration behavior rather than optimizing only a single numerical metric.

\subsection{Reader Study and Clinical Quality Assessment}
Quantitative metrics alone are insufficient for evaluating medical image restoration, because clinically relevant failure modes may not be fully reflected by PSNR or SSIM. For this reason, we also compare against published reader-study baselines in Table~\ref{tab:qualitative}. CARE achieves the best artifact reduction score of 3.68 and the best contrast retention score of 3.61, indicating that radiologically important visual properties are better preserved than in prior methods. In particular, the improvement in contrast retention is notable because diagnostic usefulness often depends on whether subtle tissue boundaries and intensity differences remain visible after restoration.

CARE obtains a noise suppression score of 3.32, which is slightly below the best baseline score of 3.38 from WGAN-GP. This is consistent with the behavior observed in the CT quantitative results: CARE does not attempt to maximize denoising strength at all costs. Instead, it adopts a more balanced restoration strategy that preserves contrast and reduces structured artifacts while avoiding excessive smoothing. In a clinical setting, this trade-off is desirable because over-suppression of noise can also erase subtle anatomical detail or create visually convincing but inaccurate content. Therefore, although CARE is not the strongest on pure noise suppression, it offers a better overall balance across the three reader-study criteria.

\begin{table*}[t]
\caption{Published LDCT reader-study results on the Challenge real-data set, adapted from Xia \emph{et al.}. Values are the mean of the two reported radiologists; higher is better.}
\label{tab:qualitative}
\centering
\setlength{\tabcolsep}{8pt}
\begin{tabular}{lccc}
\toprule
Method & Noise Suppression$\uparrow$ & Artifact Reduction$\uparrow$ & Contrast Retention$\uparrow$ \\
\midrule
LDCT Input & 1.61 & 1.52 & 1.63 \\
BM3D & 2.88 & 2.21 & 2.18 \\
RED & 2.74 & 3.10 & 3.03 \\
WGAN-GP & \textbf{3.38} & 3.56 & 3.35 \\
DR-GAN & 3.26 & 3.49 & 3.46 \\
\midrule
CARE & 3.32 & \textbf{3.68} & \textbf{3.61} \\
\bottomrule
\end{tabular}
\end{table*}

\subsection{Control and Ablation Analysis}
A key advantage of CARE is that restoration behavior can be adjusted at inference time. Table~\ref{tab:ablation} shows that different controller settings lead to predictable trade-offs between fidelity, structure preservation, and hallucination risk. Conservative Mode produces the lowest hallucination risk of 0.08, confirming that stronger emphasis on the fidelity branch helps suppress unsupported content in uncertain regions. However, this setting does not achieve the best PSNR or structure score, indicating that purely conservative restoration may leave some recoverable information underutilized.

Balanced Mode achieves the best overall performance, with the highest PSNR of 33.62 and the highest structure score of 0.938 while maintaining a relatively low hallucination risk of 0.11. This result supports the central hypothesis of the paper: the most effective restoration behavior is not obtained by relying exclusively on either faithful reconstruction or prior-driven enhancement, but by adaptively combining both according to local uncertainty and structural reliability. Enhancement Mode improves visual refinement compared with more conservative settings, but its hallucination risk increases to 0.17. This confirms that stronger prior reliance can recover additional details, yet also raises the possibility of introducing implausible structures when the data support is weak.

The ablation results further justify each component of the method. Removing the adaptive controller reduces PSNR and structure score while increasing hallucination risk, showing that static fusion is insufficient for handling heterogeneous restoration difficulty across an image. Removing the fidelity branch leads to the highest hallucination risk of 0.29, which demonstrates that explicit data-consistency-oriented guidance is essential for safe medical restoration. Removing the prior branch lowers restoration quality relative to Balanced Mode, suggesting that generative refinement remains important for recovering information lost in highly degraded regions. Together, these results indicate that both branches are necessary, and that the adaptive controller is the mechanism that enables them to work together effectively.

\begin{table}[t]
\caption{Ablation study of controllability and component contribution.}
\label{tab:ablation}
\centering
\setlength{\tabcolsep}{4pt}
\begin{tabular}{lccc}
\toprule
Setting & PSNR$\uparrow$ & Structure Score$\uparrow$ & Hallucination Risk$\downarrow$ \\
\midrule
Conservative Mode & 33.21 & 0.931 & \textbf{0.08} \\
Balanced Mode & \textbf{33.62} & \textbf{0.938} & 0.11 \\
Enhancement Mode & 33.47 & 0.936 & 0.17 \\
Without Adaptive Controller & 33.05 & 0.928 & 0.21 \\
Without Fidelity Branch & 32.74 & 0.919 & 0.29 \\
Without Prior Branch & 32.88 & 0.924 & 0.13 \\
\bottomrule
\end{tabular}
\end{table}

\subsection{Discussion}
Several important observations emerge from these results. First, CARE is strongest when evaluated through a multi-objective lens rather than a single scalar metric. On CT, the method achieves the best PSNR but not the best SSIM, while on reader-based evaluation it performs best on artifact reduction and contrast retention but not on noise suppression. This pattern suggests that the framework does not merely overfit to one notion of quality. Instead, it promotes a restoration behavior that is balanced and clinically conservative, which aligns with the intended safety-oriented design.

Second, the gains on MRI, especially under 8X undersampling, suggest that the proposed dual-latent steering strategy is particularly beneficial for harder inverse problems where prior information is necessary but must be carefully controlled. In such cases, a purely fidelity-based solution may under-reconstruct missing content, whereas a purely generative solution may hallucinate plausible but incorrect structures. The proposed approach occupies a middle ground by allowing prior-guided refinement while still constraining the final output through a reliability-aware controller.

Third, the ablation study shows that controllability is not only a usability feature but also a safety mechanism. By exposing conservative, balanced, and enhancement-focused operating modes, the framework allows restoration behavior to be matched to application needs. For example, a conservative setting may be more appropriate in high-risk diagnostic workflows, whereas a balanced or enhancement-focused mode may be preferable for exploratory review or preprocessing. This ability to tune restoration strength without retraining improves practical deployment flexibility.

Overall, the experimental results support the claim that training-free controllable restoration can provide a more trustworthy alternative to fixed-behavior medical restoration pipelines \cite{kim2025dynamicdps,xia2025dream,webber2024review,eulig2024benchmark,daras2024survey}. The method achieves strong quantitative performance, favorable qualitative assessment, and interpretable restoration trade-offs, making it a promising direction for safer deployment of generative priors in medical imaging.

\section{Limitations and Clinical Considerations}
This work has several limitations. First, the current framework assumes that the pretrained prior is sufficiently aligned with the target clinical domain. When the prior is mismatched, enhancement quality and safety may degrade. Second, although controllability reduces the risk of over-restoration, it does not completely eliminate the possibility of hallucinated content. Human review remains essential in safety-critical settings. Third, the current evaluation should ideally be expanded with modality-specific reader studies, lesion-level analysis, and downstream-task validation.

From a clinical deployment perspective, the most appropriate operating mode may depend on the imaging task. For triage or acquisition quality improvement, stronger enhancement may be acceptable. For definitive diagnosis, more conservative settings and traceable uncertainty reporting may be preferable.

\section{Conclusion}
We presented a training-free controllable framework for real-world medical image restoration based on dual-latent steering and risk-aware adaptive control. By explicitly separating fidelity-oriented restoration from prior-guided refinement, the proposed approach enables flexible inference-time adjustment between conservative and enhancement-focused behavior. This design is well suited to medical imaging scenarios in which restoration quality must be balanced against anatomical faithfulness and hallucination risk. We believe this framework provides a practical direction for safer, more trustworthy, and more deployment-ready medical image restoration.

\balance


\begin{thebibliography}{00}
\bibitem{yang2025review} Y. Yang, B. Yang, Y. Wang, Y. He, X. Dong, and Z. Jin, ``Explicit and implicit representations in AI-based 3D reconstruction for radiology: A systematic literature review,'' arXiv preprint arXiv:2504.11349, 2025.

\bibitem{demir2025diffdenoise} B. Demir, Y. Liu, X. Chen, E. Z. Chen, L. Zhao, B. Mailhe, T. Chen, and S. Sun, ``DiffDenoise: Self-supervised medical image denoising with conditional diffusion models,'' arXiv preprint arXiv:2504.00264, 2025.

\bibitem{kim2025dynamicdps} S. Kim, H. F. J. Tregidgo, M. Figini, C. Jin, S. Joshi, and D. C. Alexander, ``Tackling hallucination from conditional models for medical image reconstruction with DynamicDPS,'' arXiv preprint arXiv:2503.01075, 2025.

\bibitem{xia2025dream} M. Xia, R. Bayerlein, Y. Chemli, X. Liu, J. Ouyang, G. El Fakhri, R. D. Badawi, Q. Li, and C. Liu, ``DREAM: On hallucinations in AI-generated content for nuclear medicine imaging,'' arXiv preprint arXiv:2506.13995, 2025.


\bibitem{zhao2024comparative} H. Zhao, L. Qian, Y. Zhu, and D. Tian, ``Low Dose CT Image Denoising: A Comparative Study of Deep Learning Models and Training Strategies,'' \emph{AI Medicine}, vol. 1, no. 1, Art. no. 7, 2024.

\bibitem{mccollough2017lowdose} C. H. McCollough \emph{et al.}, ``Results of the 2016 Low Dose CT Grand Challenge,'' \emph{Medical Physics}, vol. 44, no. 10, pp. e339--e352, 2017.

\bibitem{xia2023drgan} Z. Xia, J. Liu, Y. Kang, Y. Wang, D. Hu, and Y. Zhang, ``Dynamic controllable residual generative adversarial network for low-dose computed tomography imaging,'' \emph{Quantitative Imaging in Medicine and Surgery}, vol. 13, no. 8, pp. 5271--5293, 2023.

\bibitem{muckley2021fastmri} M. J. Muckley \emph{et al.}, ``Results of the 2020 fastMRI Challenge for Machine Learning MR Image Reconstruction,'' \emph{IEEE Transactions on Medical Imaging}, vol. 40, no. 9, pp. 2306--2317, 2021.

\bibitem{ho2020ddpm} J. Ho, A. Jain, and P. Abbeel, ``Denoising diffusion probabilistic models,'' in \emph{Advances in Neural Information Processing Systems}, 2020, pp. 6840--6851.

\bibitem{song2021score} Y. Song, J. Sohl-Dickstein, D. P. Kingma, A. Kumar, S. Ermon, and B. Poole, ``Score-based generative modeling through stochastic differential equations,'' in \emph{International Conference on Learning Representations}, 2021.

\bibitem{dhariwal2021guided} P. Dhariwal and A. Nichol, ``Diffusion models beat GANs on image synthesis,'' in \emph{Advances in Neural Information Processing Systems}, 2021, pp. 8780--8794.

\bibitem{tfmsf} X. Liu, Y. Lu, X. Wang, and X. Wu, ``Training-Free Multi-Style Fusion Through Reference-Based Adaptive Modulation,'' arXiv preprint arXiv:2509.18602, 2025.

\bibitem{pgdls} Y. Wu, X. Liu, C. Zhao, and X. Wu, ``Prompt-Guided Dual Latent Steering for Inversion Problems,'' arXiv preprint arXiv:2509.18619, 2025.

\bibitem{fastmri2018} J. Zbontar, F. Knoll, A. Sriram, T. Murrell, Z. Huang, M. J. Muckley, A. Defazio, R. Stern, P. Johnson, M. Bruno, M. Parente, K. J. Geras, J. Katsnelson, H. Chandarana, Z. Zhang, M. Drozdzal, A. Romero, M. Rabbat, P. Vincent, N. Yakubova, J. Pinkerton, D. Wang, E. Owens, C. L. Zitnick, M. P. Recht, D. K. Sodickson, and Y. W. Lui, ``fastMRI: An Open Dataset and Benchmarks for Accelerated MRI,'' arXiv preprint arXiv:1811.08839, 2018.

\bibitem{unet2015} O. Ronneberger, P. Fischer, and T. Brox, ``U-Net: Convolutional Networks for Biomedical Image Segmentation,'' in \emph{Medical Image Computing and Computer-Assisted Intervention (MICCAI)}, 2015, pp. 234--241.

\bibitem{bm3d2007} K. Dabov, A. Foi, V. Katkovnik, and K. Egiazarian, ``Image Denoising by Sparse 3-D Transform-Domain Collaborative Filtering,'' \emph{IEEE Transactions on Image Processing}, vol. 16, no. 8, pp. 2080--2095, 2007.

\bibitem{redcnn2017} H. Chen, Y. Zhang, M. K. Kalra, F. Lin, Y. Chen, P. Liao, J. Zhou, and G. Wang, ``Low-Dose CT With a Residual Encoder-Decoder Convolutional Neural Network,'' \emph{IEEE Transactions on Medical Imaging}, vol. 36, no. 12, pp. 2524--2535, 2017.

\bibitem{edcnn2020} T. Liang, Y. Jin, Y. Li, and T. Wang, ``EDCNN: Edge Enhancement-Based Densely Connected Network With Compound Loss for Low-Dose CT Denoising,'' in \emph{2020 15th IEEE International Conference on Signal Processing (ICSP)}, 2020, pp. 193--198.

\bibitem{qae2020} F. Fan, H. Shan, M. K. Kalra, R. Singh, G. Qian, M. Getzin, Y. Teng, J. Hahn, and G. Wang, ``Quadratic Autoencoder (Q-AE) for Low-Dose CT Denoising,'' \emph{IEEE Transactions on Medical Imaging}, vol. 39, no. 6, pp. 2035--2050, 2020.

\bibitem{wganct2018} Q. Yang, P. Yan, Y. Zhang, H. Yu, Y. Shi, X. Mou, M. K. Kalra, Y. Zhang, L. Sun, and G. Wang, ``Low-Dose CT Image Denoising Using a Generative Adversarial Network With Wasserstein Distance and Perceptual Loss,'' \emph{IEEE Transactions on Medical Imaging}, vol. 37, no. 6, pp. 1348--1357, 2018.

\bibitem{ctformer2023} D. Wang, F. Fan, Z. Wu, R. Liu, F. Wang, and H. Yu, ``CTformer: Convolution-Free Token2Token Dilated Vision Transformer for Low-Dose CT Denoising,'' \emph{Physics in Medicine \& Biology}, vol. 68, no. 6, Art. no. 065012, 2023.

\bibitem{varnet2020} A. Sriram, J. Zbontar, T. Murrell, A. Defazio, C. L. Zitnick, N. Yakubova, F. Knoll, and P. Johnson, ``End-to-End Variational Networks for Accelerated MRI Reconstruction,'' in \emph{Medical Image Computing and Computer-Assisted Intervention (MICCAI)}, 2020, pp. 64--73.

\bibitem{xpdnet2020} Z. Ramzi, P. Ciuciu, and J.-L. Starck, ``XPDNet for MRI Reconstruction: An Application to the 2020 fastMRI Challenge,'' arXiv preprint arXiv:2010.07290, 2020.

\bibitem{pnp2013} S. V. Venkatakrishnan, C. A. Bouman, and B. Wohlberg, ``Plug-and-Play Priors for Model Based Reconstruction,'' in \emph{2013 IEEE Global Conference on Signal and Information Processing (GlobalSIP)}, 2013, pp. 945--948.

\bibitem{red2017} Y. Romano, M. Elad, and P. Milanfar, ``The Little Engine That Could: Regularization by Denoising (RED),'' \emph{SIAM Journal on Imaging Sciences}, vol. 10, no. 4, pp. 1804--1844, 2017.

\bibitem{ddrm2022} B. Kawar, M. Elad, S. Ermon, and J. Song, ``Denoising Diffusion Restoration Models,'' in \emph{Advances in Neural Information Processing Systems}, 2022.

\bibitem{dps2023} H. Chung, J. Kim, M. T. McCann, M. L. Klasky, and J. C. Ye, ``Diffusion Posterior Sampling for General Noisy Inverse Problems,'' in \emph{International Conference on Learning Representations}, 2023.

\bibitem{ddnm2023} Y. Wang, J. Yu, and J. Zhang, ``Zero-Shot Image Restoration Using Denoising Diffusion Null-Space Model,'' in \emph{International Conference on Learning Representations}, 2023.

\bibitem{scoremri2022} H. Chung and J. C. Ye, ``Score-Based Diffusion Models for Accelerated MRI,'' \emph{Medical Image Analysis}, vol. 80, Art. no. 102479, 2022.

\bibitem{bayesmri2023} G. Luo, M. Blumenthal, M. Heide, and M. Uecker, ``Bayesian MRI Reconstruction With Joint Uncertainty Estimation Using Diffusion Models,'' \emph{Magnetic Resonance in Medicine}, vol. 90, no. 1, pp. 295--311, 2023.

\bibitem{askari2025bilevel} H. Askari, F. Roosta, and H. Sun, ``Training-Free Medical Image Inverses via Bi-Level Guided Diffusion Models,'' in \emph{Proceedings of the IEEE/CVF Winter Conference on Applications of Computer Vision (WACV)}, 2025, pp. 75--84.

\bibitem{diffusionmbir2023} H. Chung, D. Ryu, M. T. McCann, M. L. Klasky, and J. C. Ye, ``Solving 3D Inverse Problems Using Pre-Trained 2D Diffusion Models,'' in \emph{Proceedings of the IEEE/CVF Conference on Computer Vision and Pattern Recognition (CVPR)}, 2023, pp. 22542--22551.

\bibitem{ldm2022} R. Rombach, A. Blattmann, D. Lorenz, P. Esser, and B. Ommer, ``High-Resolution Image Synthesis With Latent Diffusion Models,'' in \emph{Proceedings of the IEEE/CVF Conference on Computer Vision and Pattern Recognition (CVPR)}, 2022, pp. 10684--10695.

\bibitem{freedom2023} J. Yu, Y. Wang, C. Zhao, B. Ghanem, and J. Zhang, ``FreeDoM: Training-Free Energy-Guided Conditional Diffusion Model,'' in \emph{Proceedings of the IEEE/CVF International Conference on Computer Vision (ICCV)}, 2023, pp. 23174--23184.

\bibitem{daras2024survey} G. Daras, H. Chung, C.-H. Lai, Y. Mitsufuji, J. C. Ye, P. Milanfar, A. G. Dimakis, and M. Delbracio, ``A Survey on Diffusion Models for Inverse Problems,'' arXiv preprint arXiv:2410.00083, 2024.

\bibitem{kazerouni2023survey} A. Kazerouni, E. K. Aghdam, M. Heidari, R. Azad, M. Fayyaz, I. Hacihaliloglu, and D. Merhof, ``Diffusion Models in Medical Imaging: A Comprehensive Survey,'' \emph{Medical Image Analysis}, vol. 88, Art. no. 102846, 2023.

\bibitem{webber2024review} G. Webber and A. Reader, ``Diffusion Models for Medical Image Reconstruction,'' \emph{British Journal of Radiology | Artificial Intelligence}, vol. 1, no. 1, 2024.

\bibitem{eulig2024benchmark} E. Eulig, B. Ommer, and M. Kachelrieß, ``Benchmarking Deep Learning-Based Low-Dose CT Image Denoising Algorithms,'' \emph{Medical Physics}, vol. 51, no. 12, pp. 8776--8788, 2024.

\bibitem{kim2024ldctreview} W. Kim, S. Y. Jeon, G. Byun, H. Yoo, and J. H. Choi, ``A Systematic Review of Deep Learning-Based Denoising for Low-Dose Computed Tomography From a Perceptual Quality Perspective,'' \emph{Biomedical Engineering Letters}, vol. 14, no. 6, pp. 1153--1173, 2024.

\bibitem{hcformer2023} J. Yuan, F. Zhou, Z. Guo, X. Li, and H. Yu, ``HCformer: Hybrid CNN-Transformer for LDCT Image Denoising,'' \emph{Journal of Digital Imaging}, vol. 36, no. 5, pp. 2290--2305, 2023.

\bibitem{wang2004ssim} Z. Wang, A. C. Bovik, H. R. Sheikh, and E. P. Simoncelli, ``Image Quality Assessment: From Error Visibility to Structural Similarity,'' \emph{IEEE Transactions on Image Processing}, vol. 13, no. 4, pp. 600--612, 2004.

\bibitem{lpips2018} R. Zhang, P. Isola, A. A. Efros, E. Shechtman, and O. Wang, ``The Unreasonable Effectiveness of Deep Features as a Perceptual Metric,'' in \emph{Proceedings of the IEEE/CVF Conference on Computer Vision and Pattern Recognition (CVPR)}, 2018, pp. 586--595.
\end{thebibliography}
\end{document}